\documentclass[nohyperref]{article}

\usepackage{microtype}
\usepackage{graphicx}
\usepackage{subcaption}
\usepackage{booktabs} %
\usepackage{siunitx}
\usepackage{dblfloatfix}

\usepackage{hyperref}

\usepackage[accepted]{icml2022}

\usepackage{amsmath}
\usepackage{amssymb}
\usepackage{mathtools}
\usepackage{amsthm}

\usepackage[capitalize,noabbrev]{cleveref}

\usepackage{comment}

\theoremstyle{plain}

\theoremstyle{definition}

\theoremstyle{remark}

\usepackage[textsize=tiny]{todonotes}

\icmltitlerunning{Self-supervised models of audio effectively explain human cortical responses to speech}

\begin{document}

\twocolumn[
\icmltitle{Self-supervised models of audio effectively explain \\ human cortical responses to speech}

\begin{icmlauthorlist}
\icmlauthor{Aditya R. Vaidya}{ut_cs}
\icmlauthor{Shailee Jain}{ut_cs}
\icmlauthor{Alexander G. Huth}{ut_cs}
\end{icmlauthorlist}

\icmlaffiliation{ut_cs}{Department of Computer Science, The University of Texas at Austin}

\icmlcorrespondingauthor{Aditya Vaidya}{avaidya@utexas.edu}

\icmlkeywords{Machine Learning, ICML} %

\vskip 0.3in
]

\begin{abstract}
Self-supervised language models are very effective at predicting high-level cortical responses during language comprehension.
However, the best current models of lower-level auditory processing in the human brain rely on either hand-constructed acoustic filters or representations from supervised audio neural networks.
In this work, we capitalize on the progress of self-supervised speech representation learning (SSL) to create new state-of-the-art models of the human auditory system.
Compared against acoustic baselines, phonemic features, and supervised models, representations from the middle layers of self-supervised models (APC, wav2vec, wav2vec 2.0, and HuBERT) consistently yield the best prediction performance for fMRI recordings within the auditory cortex (AC).
Brain areas involved in low-level auditory processing exhibit a preference for earlier SSL model layers, whereas higher-level semantic areas prefer later layers.
We show that these trends are due to the models' ability to encode information at multiple linguistic levels (acoustic, phonetic, and lexical) along their representation depth.
Overall, these results show that self-supervised models effectively capture the hierarchy of information relevant to different stages of speech processing in human cortex.
\end{abstract}

\begin{figure*}[tb]
    \centering
    \includegraphics[scale=0.36]{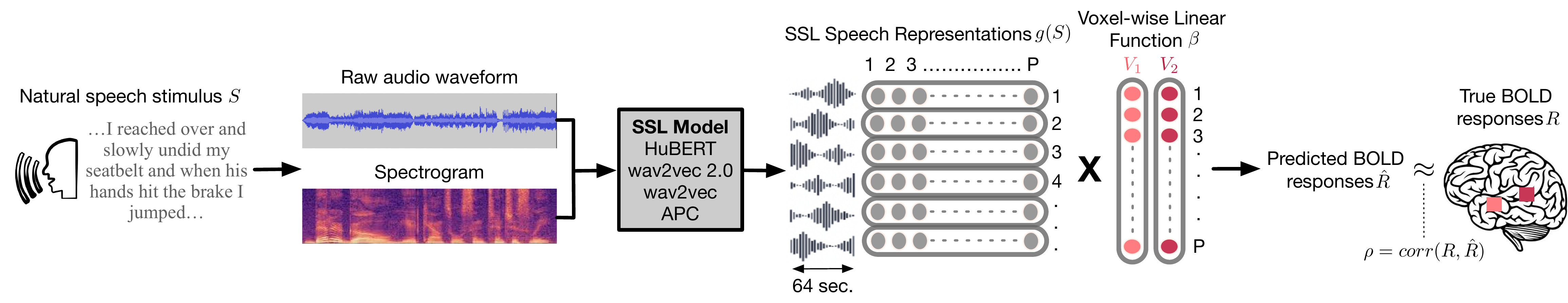}
    \caption{Voxel-wise encoding models from SSL models of speech. 64 second spans of narrative stimuli (represented as waveforms or spectrograms) are fed into an SSL model trained to learn the statistics of speech. Hidden states from a layer in the neural network are extracted to form representation $g(S)$ after downsampling to the rate of fMRI acquisition. $g(S)$ is then used to fit encoding models that predict fMRI BOLD responses to natural speech. Encoding model weights $\beta \in \mathbb{R}^{P \times V}$ are estimated with ridge regression. Models are then used to predict $\hat{R}$, the BOLD response to an unseen stimulus, and are evaluated by the correlation between $R$ and $\hat{R}$.}
    \label{fig:intro}
\end{figure*}

\section{Introduction}
Self-supervised learning (SSL) has emerged as a popular and successful pre-training objective. In natural language processing, self-supervised language models (LM) encode diverse linguistic information and achieve excellent zero-shot performance on many language tasks \cite{petersDeepContextualizedWord2018,radfordImprovingLanguageUnderstanding2018,devlinBERTPretrainingDeep2019}. Capitalizing on these findings, neuroimaging studies have shown that representations extracted from LMs are highly effective at predicting brain activity elicited by natural language \cite{wehbeAligningContextbasedStatistical2014,jainIncorporatingContextLanguage2018,tonevaInterpretingImprovingNaturallanguage2019,schrimpfNeuralArchitectureLanguage2021,caucheteuxGPT2ActivationsPredict2021,goldsteinThinkingAheadSpontaneous2020} and can help reveal how linguistic representations are organized across human cortex \cite{jainInterpretableMultitimescaleModels2020}.

In automatic speech recognition there has been a similar trend towards SSL-based approaches. Using autoregressive, contrastive, and masked prediction losses, SSL models learn powerful acoustic representations that are highly transferable across applications~\cite{hsuHuBERTSelfSupervisedSpeech2021, schneiderWav2vecUnsupervisedPretraining2019,baevskiWav2vecFrameworkSelfSupervised2020}. Much like language models, SSL models of speech have also been shown to capture higher-order linguistic structure without explicit supervision~\cite{pasadLayerwiseAnalysisSelfsupervised2021, yangSUPERBSpeechProcessing2021}. This can be attributed to their ability to integrate information over a broader context than traditional acoustic filters. However, SSL models have not yet been combined with neuroimaging to study auditory processing in the human brain, where the best models are currently hand-constructed features~\cite{norman-haignereNeuralResponsesNatural2018,veneziaHierarchySpeechdrivenSpectrotemporal2019,chiMultiresolutionSpectrotemporalAnalysis2005,mesgaraniPhoneticFeatureEncoding2014} or supervised neural networks~\cite{milletInductiveBiasesPretraining2021}.
In this paper, we investigate the potential of SSL speech representations for predicting human cortical responses to natural speech.

One approach to studying sensory representations in the brain is through ``encoding models'' \cite{wuCompleteFunctionalCharacterization2006}. These predictive models of brain activity learn a mapping from stimulus features to responses measured using a neuroimaging technique like fMRI. In this work, we use SSL models to extract acoustic features of natural speech and in turn, build voxel-wise speech encoding models using data from an fMRI experiment (\cref{fig:intro}). Our experiment comprised human participants passively listening to English-language narrative stories while their whole-brain fMRI BOLD activity was being recorded. Prior work on LM-based language encoding models has found performance differences across LM layers \cite{jainIncorporatingContextLanguage2018, tonevaInterpretingImprovingNaturallanguage2019}, and layer-wise analyses of SSL models have shown that they capture different types of acoustic information \cite{pasadLayerwiseAnalysisSelfsupervised2021}. Motivated by these findings, we built separate encoding models for each layer in four SSL models --- APC, wav2vec, wav2vec 2.0 and HuBERT (\cref{table:model-config}). Their performance was compared to low-level acoustic baselines like Mel spectrograms and spectrotemporal features, mid-level phoneme articulations, and high-level semantic features.

Overall, we found that SSL models are highly effective at predicting cortical responses to natural speech, beating supervised and hand-engineered acoustic representations. 
Despite being trained on audiobooks, the SSL representations transferred well to encoding models of natural speech. %
We further show that encoding performance greatly varied across SSL model layers and brain areas. While the upper-middle layers of some models had the best performance broadly across cortex, lower layers were comparable only in low-level primary auditory cortex. To better understand why these models worked so well, we did variance partitioning across pairs of feature spaces. 
This revealed that the best SSL layers capture spectral, phonemic, and semantic information.
Finally, we directly assessed linguistic representations in each SSL layer using linear probes.
This showed that the evolution of representations across SSL layers mirrors the presumed stages of speech processing.
Together with the encoding model results, this shows that SSL models capture diverse acoustic information across their representational depth which enables them to effectively model different stages of speech processing in cortex.

\section{Natural language fMRI experiment}
\label{sec:fmri-experiment}
To build SSL encoding models, we used data from an fMRI experiment comprising 6 participants (2 female) listening to over five hours of spoken narrative stories (27 stories; $\sim$57,900 total words) from \textit{The Moth Radio Hour}.
These rich, diverse naturalistic stimuli are highly representative of speech humans encounter on a daily basis. To find the exact timing of each word and phoneme, the stories were transcribed and a forced aligner was used to align each transcript to its audio.
All subjects were healthy with normal hearing, had fluent English language comprehension, and gave written informed consent.
Whole-brain MRI data was collected every 2 seconds (TR).
The experimental procedure was approved by the local Institutional Review Board.
More MRI acquisition details can be found in Appendix~\ref{sec:supp_mri-acquisition}.

\begin{table*}[tb]
\caption{Model architectures and training objectives. For each model, its input representation is fed into an encoder module to produce latent speech representations. A second module then produces contextualized representations that capture information across the input. Both modules are trained end-to-end to fulfill their supervised or self-supervised objective.}
\label{table:model-config}
\begin{center}
\begin{small}

\begin{tabular}{@{}lcccc@{}}
\toprule
Model                     & Input           & Encoder     & Contextualizer       & Loss type          \\ \midrule
APC~\cite{chungUnsupervisedAutoregressiveModel2019, chungGenerativePreTrainingSpeech2020}                       & log Mel spectrogram & N/A         & 3-layer GRU          & Autoregressive     \\
wav2vec~\cite{schneiderWav2vecUnsupervisedPretraining2019}                   & Waveform        & 7-layer CNN & 12-layer CNN         & Contrastive        \\
wav2vec 2.0 \textsc{base}~\cite{baevskiWav2vecFrameworkSelfSupervised2020} & Waveform        & 7-layer CNN & 12-layer Transformer & Masked contrastive        \\
HuBERT \textsc{base}~\cite{hsuHuBERTSelfSupervisedSpeech2021}      & Waveform        & 7-layer CNN & 12-layer Transformer & Masked predictive \\
Deep Speech 2~\cite{amodeiDeepSpeechEndtoend2016}             & log spectrogram & 2-layer CNN & 5-layer LSTM         & CTC (supervised)   \\ \bottomrule
\end{tabular}

\end{small}
\end{center}
\end{table*}

\section{Voxel-wise encoding models}
Encoding models aim to approximate $f(S)=R$, the mapping between a stimulus $S$ and the elicited BOLD response $R$ measured in some brain area (\cref{fig:intro}). Here, the stimulus $S$ can take any form, like images, audio waveforms, or words. To make model fitting tractable with limited fMRI data, we restricted encoding models to the \textit{linearized} form $f(S)=g(S) \beta$, where $g$ is a pre-specified non-linear transformation on the stimulus and $\beta$ is a vector of learned linear weights~\cite{wuCompleteFunctionalCharacterization2006}.
If $g$ transforms $S$ into a $P$-dimensional feature space and $V$ is the number of cortical voxels (regression targets), then $\beta$ is a $P \times V$ linear transformation between the feature space and each voxel's predicted response.
We estimated a separate encoding model $\hat{f}_v$ for each voxel $v$ on the training dataset $(S_\text{train}, R_\text{train})$ using ridge regression. To select the ridge parameter independently for each voxel, we used 50 iterations of cross-validation. Since fMRI data is auto-correlated, for each cross-validation run we randomly sampled 40 different chunks of the training data, each totaling over 4 minutes. The training set comprised 26 stories, totaling ~5.4 hours.

For evaluation, we used the learned encoding models to predict the response timecourse of each voxel $v$ on a held-out test set, $\hat{R}_{\text{test},v} = \hat{f}_v (S_{\text{test}})$.
We then calculated the linear correlation between true and predicted responses to determine encoding performance $\rho_v = \text{corr} ( R_{\text{test},v}, \hat{R}_{\text{test},v} )$. The test set comprised 1 held-out story (10 minutes) that did not participate in model estimation.

Using the encoding model framework, we can compare how well different feature spaces model speech processing in the brain. Consider a feature space $g_i$ on which we estimate and test encoding models, yielding prediction performance $\rho_v^i$ for a voxel $v$. If $\rho_v^1 > \rho_v^2$ for some feature spaces $g_1$ and $g_2$, we can conclude that $g_1$ is a closer match to the information encoded by voxel $v$ while processing speech. By examining the type of information encoded in $g_1$, we can consequently estimate voxel function. Here we use this approach to compare representations extracted from SSL speech models with several well-known baselines and supervised alternatives.

\subsection{Extracting speech features from SSL Models}

We extracted hidden state representations from each layer of four self-supervised models: APC~\cite{chungUnsupervisedAutoregressiveModel2019,chungGenerativePreTrainingSpeech2020}, wav2vec~\cite{schneiderWav2vecUnsupervisedPretraining2019}, wav2vec 2.0~\cite{baevskiWav2vecFrameworkSelfSupervised2020}, and HuBERT~\cite{hsuHuBERTSelfSupervisedSpeech2021}.
These models are pre-trained on 960 hours of speech from LibriSpeech~\cite{panayotovLibrispeechASRCorpus2015}, a corpus of English audiobooks from the LibriVox project.
The SSL models are not finetuned on any annotated samples.
See \cref{table:model-config} for an overview of model architectures and training methods. In the main analyses, we evaluate the output of CNN encoders and all layers of the contextualizer.

The self-supervised speech models effectively capture statistical regularities in speech, either by learning to predict the content of future or unknown time blocks (autoregressive \& masked prediction objectives), or by learning features that are effective at discriminating future samples from other samples (contrastive objective). While APC operates on Mel spectrograms of the input, the other three models operate directly on the waveform.

\subsection{Supervised speech model baseline}
To compare SSL-based encoding models with related work~\cite{milletInductiveBiasesPretraining2021}, we additionally extracted representations from Deep Speech 2~\cite{amodeiDeepSpeechEndtoend2016}.
This automatic speech recognition model was trained with full supervision on the same amount of data as the self-supervised models (960 hours of LibriSpeech).

\subsection{Hand-engineered baselines}

We also compared encoding models trained with SSL representations against those trained with traditional acoustic and hand-labeled features.

We first used spectrotemporal modulations, which are a standard model of primary auditory cortex responses to sound~\cite{norman-haignereNeuralResponsesNatural2018,chiSpectrotemporalModulationTransfer1999,chiMultiresolutionSpectrotemporalAnalysis2005,veneziaHierarchySpeechdrivenSpectrotemporal2019}.
These features are computed by convolving spectrograms with filters that are selective to different rates of spectral and/or temporal modulation, such as changing tones or harmonics across time. We also computed spectral features (``FBANK'') by applying Mel-scale triangular filters to the power spectrum.

Since parts of auditory cortex are known to selectively respond to phonemes~\cite{mesgaraniPhoneticFeatureEncoding2014}, we also used phoneme articulations as a mid-level speech feature. These were derived by mapping hand-labeled phonemes onto 14 articulatory features.
Finally, we used a 985-dimensional word embedding to capture lexical and semantic information~\cite{huthNaturalSpeechReveals2016}.

Since some of the artificial neural networks discussed here are bi-directional, we enforced causality by extracting representations at the end of a sliding window of size \SI{64}{\second} with a stride of \SI{10}{\milli\second}.
Before their use in encoding models, all features were down-sampled to the rate of fMRI acquisition (\SI{0.5}{\hertz}) with Lanczos resampling. The haemodynamic response function for each encoding model was estimated using a finite impulse response model with 4 delays.

\section{Experiments}

\begin{figure*}[tb]
    \centering
    \includegraphics[width=\textwidth]{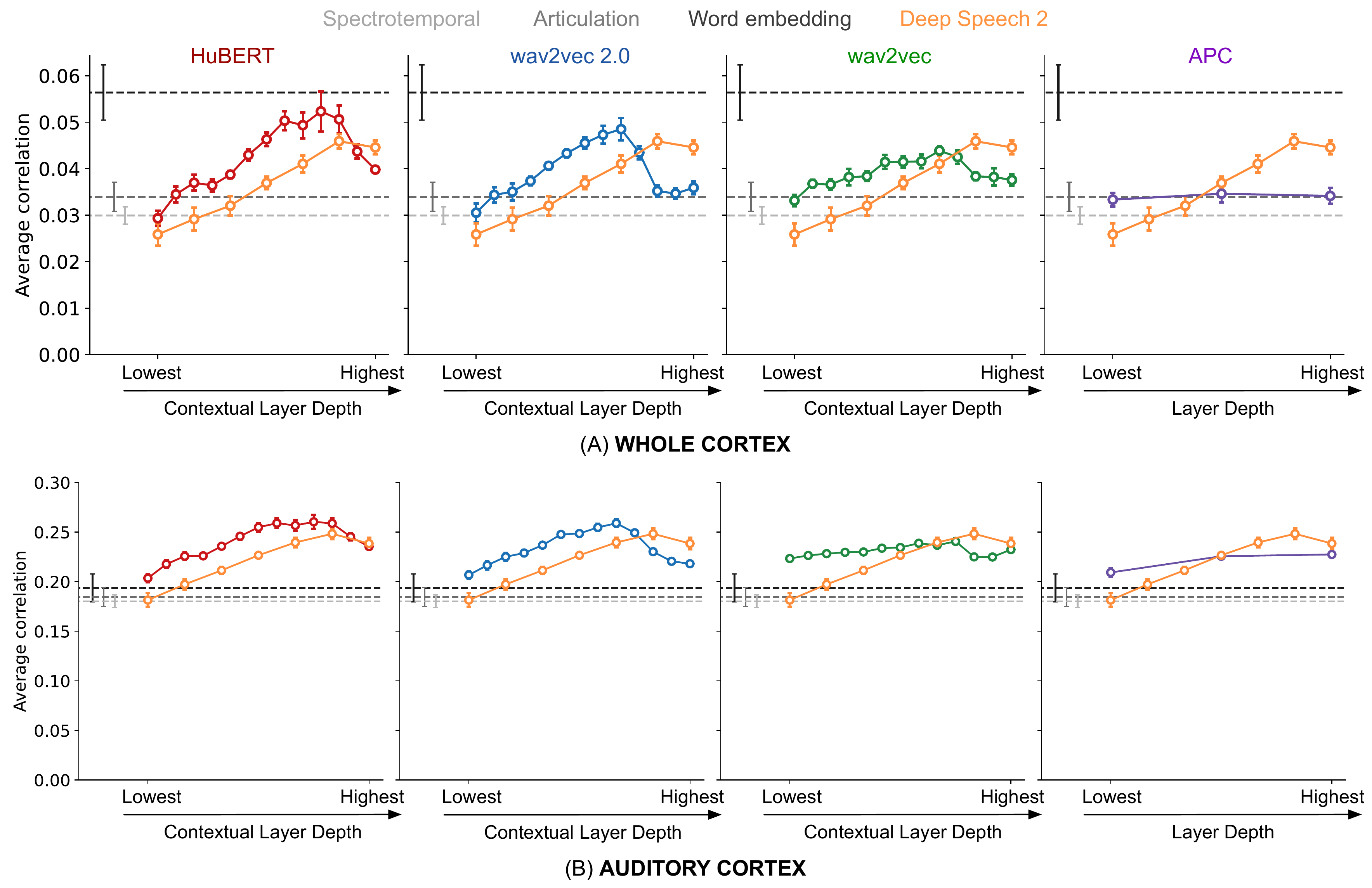}
    \caption{Encoding performance for each feature space, averaged across voxels and then subjects ($N=6$). Error bars show standard error of the mean (SEM) across subjects after correcting for per-subject overall performance (i.e., subtracting the mean performance across all models from each subject's values). 
    Each column compares one SSL model against baseline representations. (A) Encoding performance is averaged across all voxels in the cortex. HuBERT layer 9 achieves the highest performance, outperforming both auditory baselines. (B) To focus on encoding performance for lower-level features, we averaged only within auditory cortex (broadly defined). All SSL models and Deep Speech 2 outperform the hand-engineered baselines, with HuBERT still outperforming all other models. More speech ROIs can be seen in Appendix~\ref{sec:supp_rois}. The encoding model built on FBANK features had the worst performance (0.024462$\pm$0.0005) and is not visualized here.}
    \label{fig:encoding-perf-layers}
\end{figure*}

\subsection{Encoding performance comparisons}
Earlier studies have found that encoding model performance can vary widely across layers of LMs \cite{jainIncorporatingContextLanguage2018,tonevaInterpretingImprovingNaturallanguage2019} and supervised speech models \cite{milletInductiveBiasesPretraining2021,kellTaskOptimizedNeuralNetwork2018}. To test whether the same is true for SSL speech models, we measured prediction performance for encoding models built using each layer of each network shown in \cref{table:model-config}. For comparison, we also tested encoding models using each layer of the supervised speech baseline and hand-constructed feature baselines.

\cref{fig:encoding-perf-layers}A shows the average voxel prediction performance across the whole cortex for each encoding model. Across layers, this analysis reveals a consistent trend --- upper-middle layers of every model have the best encoding performance, with lower performance for the shallowest and deepest layers.
This trend is most evident in wav2vec 2.0, HuBERT, and Deep Speech 2, and is consistent with previous literature in higher-level language encoding models that also found best performance in the upper-middle layers \cite{jainIncorporatingContextLanguage2018, tonevaInterpretingImprovingNaturallanguage2019,caucheteuxGPT2ActivationsPredict2021,goldsteinThinkingAheadSpontaneous2020}.

Across models, we see that the best layer from each SSL and Deep Speech 2 layer outperform the auditory baselines (articulation and spectrotemporal modulation). Despite being trained on the same amount of data, the best layers of SSL models wav2vec 2.0 and HuBERT outperform the fully supervised Deep Speech 2 model that was used in prior work investigating speech processing across cortex \cite{milletInductiveBiasesPretraining2021}. Further, the best layer of HuBERT approaches the performance of the word embedding model, despite not receiving any explicit supervision on lexical or semantic information. 
\cref{fig:encoding-perf-layers}B shows average voxel prediction performance only for voxels within auditory cortex (AC), which is thought to be responsible for extracting higher-level features such as words from incoming acoustic information \cite{hickokCorticalOrganizationSpeech2007}. Here, most SSL layers and supervised Deep Speech 2 layers outperform all of the hand-constructed feature spaces, including word embeddings. The best encoding model overall is the same, however: layer 9 of HuBERT.

\begin{figure*}[t]
    \centering
    \includegraphics[width=\textwidth]{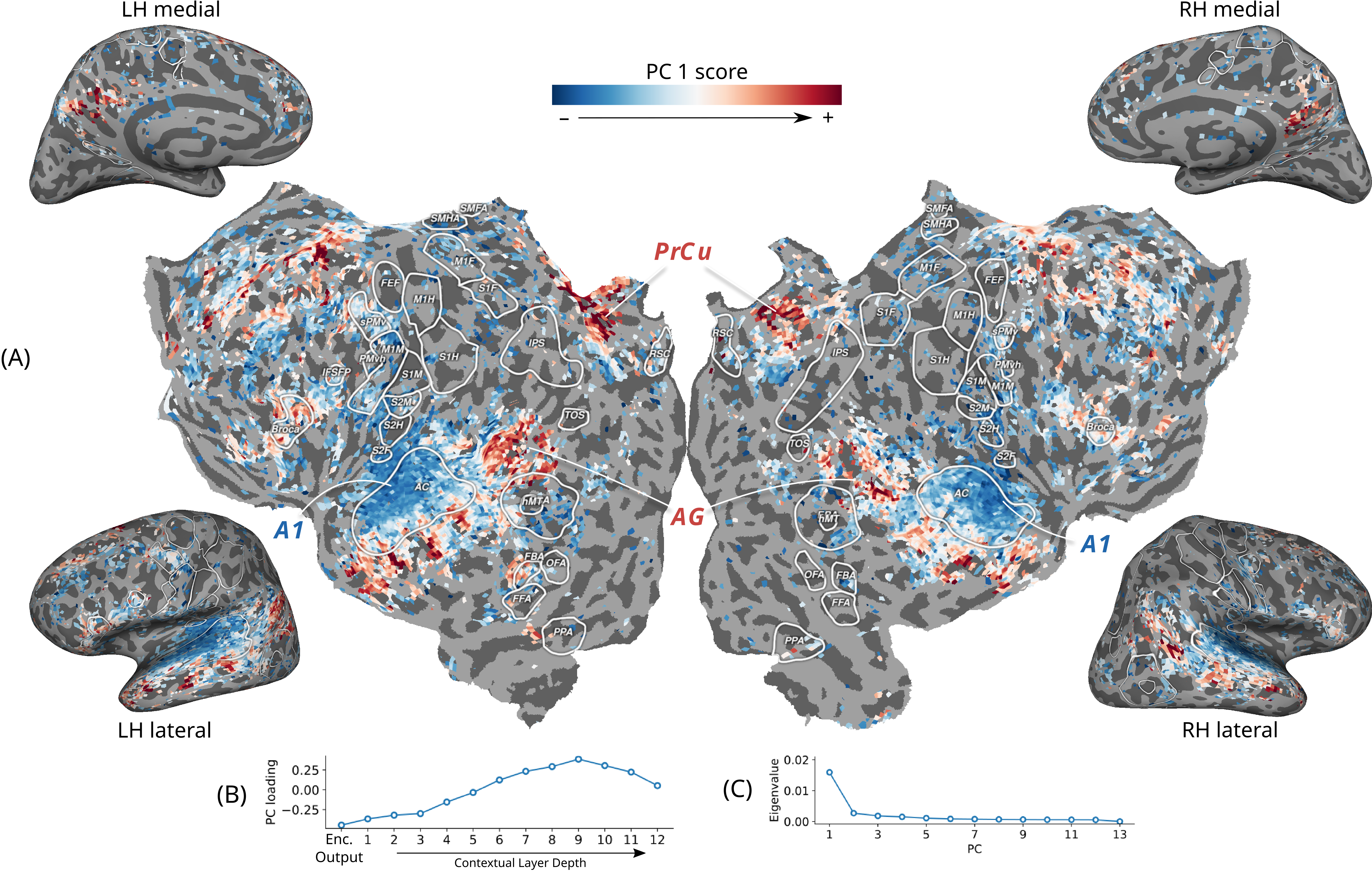}
    
    \caption{Cortical map of HuBERT layer selectivity. PCA was applied to an $n_{\text{voxels}} \times n_{\text{layers}}$ matrix $C$ containing per-voxel encoding performance for each layer in HuBERT. (A) Voxel scores for the first principal component (PC 1) across well-predicted voxels (mean $(\rho_v) > 0.15$). The cortical surface is first inflated (medial and lateral views), and then cut and flattened (flatmap, center). (B) Loadings of PC 1 indicate that the primary dimension of variance is between lower layers and upper-middle layers. Voxels with low scores (blue) prefer earlier layers, while those with high scores (red) prefer layer layers. (C) Scree plot of variance explained by each PC. We focus our analysis on the first PC because it explains much more variance than any others.}
    \label{fig:layer-selectivity}
\end{figure*}

These results could indicate that the best SSL layers are able to capture some of the same semantic information that is in word embeddings, or that the SSL models are simply better acoustic representations than the other models, or a combination of both. Our next analyses are aimed at disentangling what types of information are captured by the different SSL layers.

\subsection{Voxel-wise layer selectivity}
One way to disentangle what is represented in the SSL layers is to compare which brain areas are predicted by each layer. Earlier results provide relatively strong priors about the function of each brain area; for example, primary auditory cortex (A1) on Heschl's gyrus represents the acoustic properties of incoming sound \cite{hickokCorticalOrganizationSpeech2007}, while angular gyrus (AG) and precuneus (PrCu) are part of the brain's `semantic system' \cite{binderWhereSemanticSystem2009}. If a layer is highly predictive of A1 we may infer it contains acoustic information, and if a layer is highly predictive of AG or PrCu we may infer it contains semantic information.

Rather than examining brain maps for each layer separately, we found the major patterns of variation in prediction performance across model layers using principal components analysis (PCA). 
For an SSL model, we first construct a data matrix matrix $C$ with dimensions $V \times L$, where $V$ is the number of voxels for a subject and $L$ is the number of layers in the model.
$C_{vl}$ is then the encoding performance of layer $l$ in the SSL model for voxel $v$.
To account for overall performance differences between layers and voxels, we centered each row and column to have zero mean.
Finally, we applied principal components analysis (PCA) to $C$.

For simplicity, here we focus on the overall best-performing SSL model, HuBERT ($L=13$). The first principal component (PC) explained 57\% of the variance in $C$, while each subsequent PC explained less than 10\% of the variance (\cref{fig:layer-selectivity}C). Inspecting the first PC loadings shows that it separates voxels with high performance in the upper-middle layers from voxels with high performance in the lower layers (\cref{fig:layer-selectivity}B). We visualized layer selectivity across the cortical surface of one subject by projecting $C$ onto its first PC (\cref{fig:layer-selectivity}A).

\begin{figure*}[tb]
    \centering
    \includegraphics[width=\linewidth]{{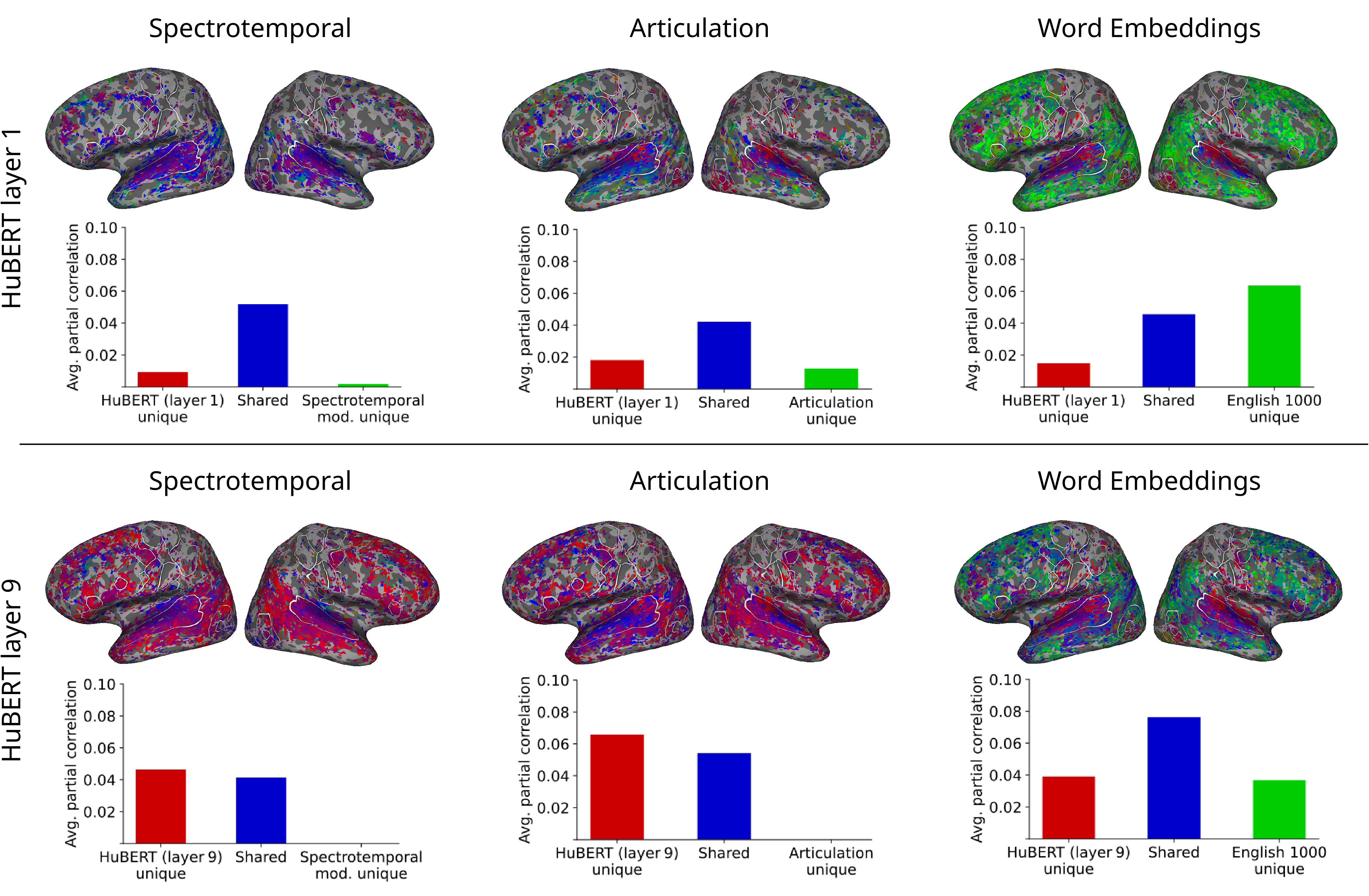}}
    \caption{Partitioning variance explained by baselines and HuBERT layers 1 and 9. For each pair of features 1 and 2, bar plots (below) show $\rho_v^{1\setminus 2}$ (red), $\rho_v^{1\cap2}$ (blue) and $\rho_v^{2\setminus 1}$ (green) averaged across cortex. Cortical maps (above) show the largest partition per voxel. Voxels are only shown if $\rho_v^{1 \cup 2} > 0.15$. Baselines are ordered by increasing level of abstraction. Every baseline has unique variance when partitioned against layer 1, whereas spectrotemporal (left) and articulatory (middle) features explain no unique variance when partitioned against layer 9. Word embeddings (right) share more variance explained with layer 9 than with layer 1.}
    \label{fig:varpart-baselines}
\end{figure*}

Overall, layer selectivity roughly follows the hierarchy of speech processing across cortex \cite{hickokCorticalOrganizationSpeech2007,binderWhereSemanticSystem2009}. While low-level regions like primary auditory cortex (A1) preferred lower HuBERT layers (negative PC 1 score), high-level regions like the angular gyrus (AG) and precuneus (PrCu) preferred the upper-middle layers (positive PC 1 score). We also compared the first PC scores to voxel encoding performance for low-level acoustic features and word embeddings. As expected from the cortical organization of PC 1 scores, low-level encoding performance was negatively correlated with PC 1 scores ($\rho=-0.330$), while word embedding encoding performance was positively correlated ($\rho=0.449$) (Appendix~\ref{sec:supp_layer-selectivity}). This suggests that the type of information varies along the depth of SSL models, with the most semantic information appearing at the upper-middle layers and most acoustic information appearing at the lowest layers. 

\subsection{Partitioning explained variance between SSL models and hand-engineered baselines}
Another way to disentangle SSL representations is to measure the overlap in brain variance explained by SSL model layers and the hand-constructed feature spaces.
To accomplish this we used variance partitioning, which separates the brain response variance that can be explained by two models into their unique and overlapping contributions \cite{deheerHierarchicalCorticalOrganization2017, lebelVoxelwiseEncodingModels2021}. 

For two feature spaces, this is done by fitting separate encoding models for each space as well as a joint encoding model, obtained by concatenating the features. 
Consider the resultant encoding performances to be $\rho_v^1$ for feature space 1, $\rho_v^2$ for feature space 2 and $\rho_v^{1\cup2}$ for the joint model. The amount of variance in the BOLD responses explained by any encoding model can be approximated as the signed squared correlation coefficient $(\rho)^2\cdot\mbox{sgn}(\rho)$, which we will denote $(\rho)^2$ for simplicity. Using set arithmetic, we can then derive the size of the intersection $(\rho^2)_v^{1\cap2} = (\rho^2)_v^1 + (\rho^2)_v^2 - (\rho^2)_v^{1\cup2}$ and $\rho_v^{1\cap2} = \sqrt{(\rho^2)_v^{1\cap2}}$. This measures the amount of BOLD response in voxel $v$ that can be equally well explained by either of the two feature spaces. Similarly, the unique contribution of feature space 1 can be computed as $(\rho^2)_v^{1\setminus 2} = (\rho^2)_v^1 - (\rho^2)_v^{1\cap2}$ with $\rho_v^{1\setminus 2} = \sqrt{(\rho^2)_v^{1\setminus 2}}$ and that of feature space 2 as $(\rho^2)_v^{2\setminus 1} = (\rho^2)_v^2 - (\rho^2)_v^{1\cap2}$ with $\rho_v^{2\setminus 1} = \sqrt{(\rho^2)_v^{2\setminus 1}}$. These values indicate variance uniquely explained by one feature space that is absent from the other.

Variance partitioning quickly becomes intractable as the number of feature spaces increases. Thus we restrict our analyses to pairwise comparisons that involve an upper-middle layer in HuBERT (layer 9) or a lower layer (layer 1). We compared each of these to the hand-constructed spectrotemporal, articulation, and word embedding feature spaces, totaling $2\times 3=6$ comparisons. For robustness, we used banded ridge regression, which allows different regularization parameters for each feature space in the joint encoding model~\cite{nunez-elizaldeVoxelwiseEncodingModels2019}. By examining the amount of unique and shared variance between each baseline and HuBERT layer, we can quantify the amount of brain-relevant acoustic, phonetic, or lexical information in each of the HuBERT layers.

\cref{fig:varpart-baselines} shows the partial correlations $\rho_v^{1\cap2}$, $\rho_v^{1\setminus 2}$ and $\rho_v^{2\setminus 1}$ for each feature space pair. Brain images (above) show the largest component of the three for each voxel across the cortical surface of one participant, and bar plots (below) show the average across the entire cortex.

We first compared the two HuBERT layers against the lowest-level baseline model, spectrotemporal modulations (\cref{fig:varpart-baselines}, column 1).
For both layers, there is a relatively large amount of shared variance and very little variance that is uniquely explained by the acoustic baseline. Layer 1 also uniquely explains very little variance, suggesting that it contains very similar information to the spectrotemporal feature space.
Layer 9, however, uniquely explains a great deal of variance over the spectrotemporal features.

Next, we compared the HuBERT layers against phoneme articulation features (\cref{fig:varpart-baselines}, column 2).
Phonemes constitute more abstract information than acoustic filters and are generally more predictive of cortical activity (\cref{fig:encoding-perf-layers}).
Here there is a larger difference between the two HuBERT layers. Layer 1 uniquely explains early auditory cortex, while articulations uniquely explain more lateral and posterior temporal cortex. 
In contrast, layer 9 appears to be a strict superset of the articulations: there is substantial shared variance, but little or no variance is uniquely explained by the articulatory model.
This layer not only captures articulatory features but also contains additional information that is not captured by those features but is relevant to the brain.

Finally, we compared the two HuBERT layers against word embeddings (\cref{fig:varpart-baselines}, column 3), which are more abstract than articulatory features.
Here, layer 9 has substantially more unique \emph{and} shared variance with word embeddings than layer 1, suggesting that the it contains more semantic information than the lower layer. Across cortex, the word embedding model uniquely explains more variance in high-level regions while layer 1 better explains AC. For layer 9, the pattern is similar but less pronounced, as most variance is shared between the two models.

Overall, these results show that the variance captured by HuBERT layer 9 entirely encompasses the low-level spectrotemporal and mid-level articulatory features, and overlaps to a great extent with the high-level word embedding features. In contrast, HuBERT layer 1 is very similar to the spectrotemporal features but fails to capture some articulatory and most semantic information. 

\subsection{Probing SSL models for linguistic structure}

\begin{figure*}[tb]
    \centering
    \includegraphics[scale=0.48]{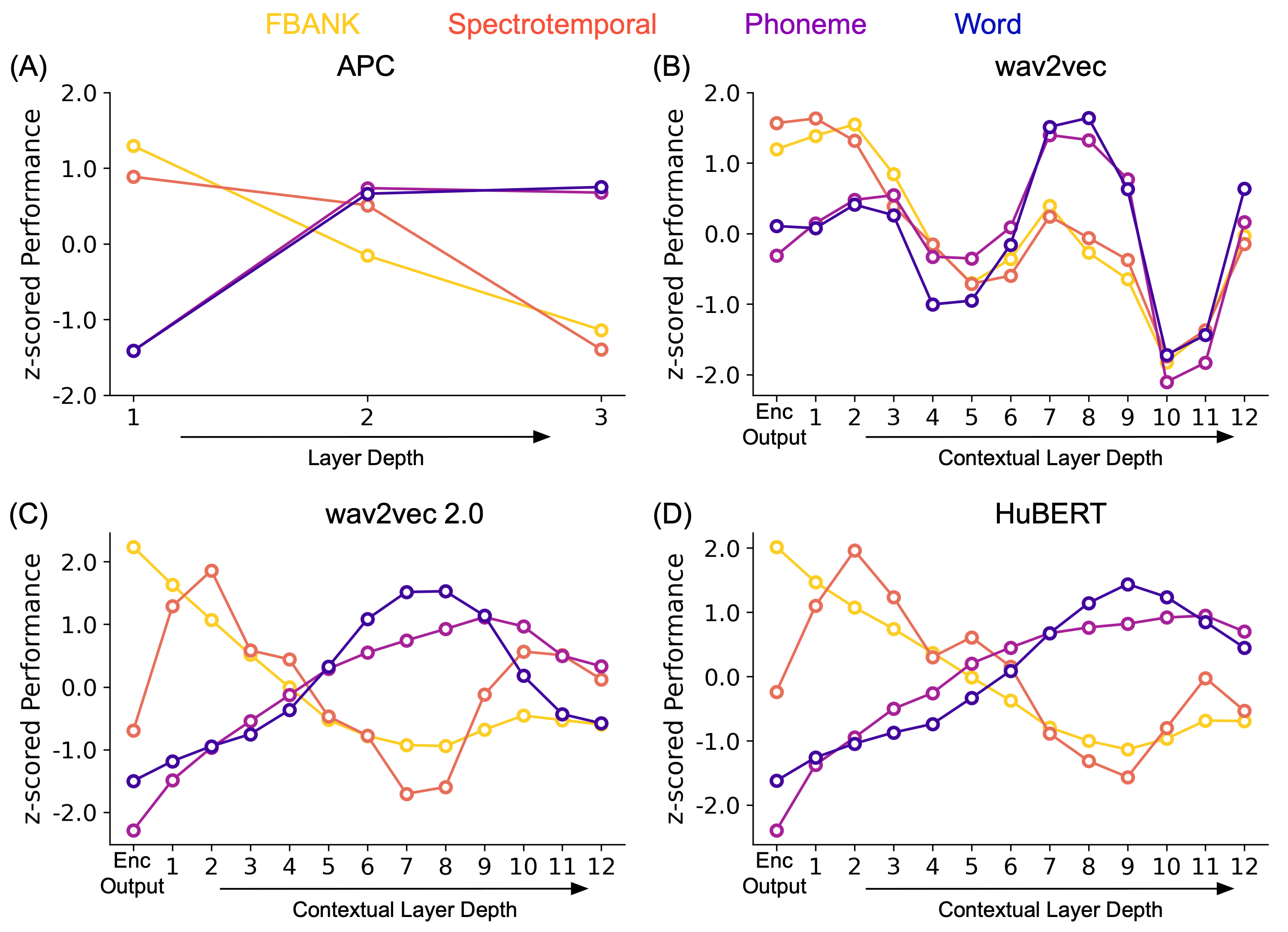}
    \caption{Probing the information represented across layers in SSL models. Each panel shows normalized performance of different layers in an SSL model predicting FBANK features, spectrotemporal features, phoneme identity, and word identity. Upper-middle layers in wav2vec, wav2vec 2.0 and HuBERT best encode phoneme- and word-level information, while the lower layers are better at predicting the acoustic features. This suggests that SSL models recapitulate the putative stages of speech processing.}
    \label{fig:probing}
\end{figure*}

The previous analyses explained much of why the SSL features are so capable at predicting brain data: the single best layer encompasses spectrotemporal, articulatory, and some semantic information. To get a finer-grained understanding of how linguistic representations evolve across these models, we turned away from the fMRI data and instead compared each layer directly to known linguistic features.
Inspired by prior work in computer vision~\cite{alainUnderstandingIntermediateLayers2017}, natural language processing~\cite{ettingerProbingSemanticEvidence2016,shiDoesStringBasedNeural2016}
and speech~\cite{pasadLayerwiseAnalysisSelfsupervised2021, yangSUPERBSpeechProcessing2021}, we did this by linearly probing each layer's representations for spectral features (FBANK), spectrotemporal features, phoneme identity, and word identity.

To build linear probes, we first obtained parallel SSL model and baseline features using the 27 fMRI stimulus stories. For baseline features with a lower sampling rate (spectrotemporal, phonemes, words), SSL layer representations were mean-pooled across time. To predict FBANK and spectrotemporal features, we used ridge regression to map each SSL model layer to each feature. Probing performance was measured by the correlation between the predicted and true feature value across the test set, averaged across all features in each space. For phoneme and word identity we trained linear classifiers and measured performance with classification accuracy and perplexity, respectively. We used 100-D GloVe embeddings \cite{penningtonGloveGlobalVectors2014} and the associated vocabulary for the word-level tasks. To ensure that these results were not biased by the limited capabilities of linear probes, we repeated some analyses using linear MLPs with a single bottleneck layer, and found no difference (Appendix~\ref{sec:supp_probing}). Each linear probe was trained and evaluated on 3 different seeds. For each run, we randomly divided the 27 stories into an 80-10-10 train-validation-test split.

\cref{fig:probing} shows the normalized probing performance of different tasks as a function of SSL model depth. 
These results suggest that the SSL layers recapitulate the putative stages of speech processing \cite{hickokCorticalOrganizationSpeech2007,liebermanUnifiedPhoneticTheory1970,pisoniStagesProcessingSpeech1975}. For most models the simplest features (FBANK) are best captured by the lowest layers, slightly more complex features (spectrotemporal) by the lower-middle layers, and high-level features (words) by the upper-middle layers. Yet phonemes, which are generally thought to fall between spectrotemporal and word-level features in speech analysis \cite{hickokCorticalOrganizationSpeech2007}, are best captured by the uppermost layers in wav2vec 2.0 and HuBERT. This may reflect how phoneme perception is influenced by word context \cite{elmanCognitivePenetrationMechanisms1988} rather than being purely based on acoustics. This effect appears to mirror human behavior despite the fact that these models have no explicit representation of words or phonemes.

\section{Conclusion}
We developed computational models of speech processing in the human cortex using representations from four different self-supervised learning (SSL) models -- APC, wav2vec, wav2vec 2.0 and HuBERT. Voxel-wise encoding models using SSL-derived features were better models of cortical activity than either phonemes or traditional acoustic features like spectrotemporal modulations and spectrograms. The best layers of wav2vec 2.0 and HuBERT also outperformed features from supervised speech models used in prior work and approached the performance of word embedding models that capture semantic information.

To better understand what information is captured by these models, we first analyzed which model layers best predicted different areas of cortex. This showed that lower layers effectively modeled areas involved in low-level acoustic processing, while the upper layers were better predictors of areas capturing phonetic and semantic information. Second, we measured the degree to which SSL layers and hand-engineered feature spaces predicted the same or different variance in brain responses. This suggested that the best SSL layer achieved its encoding performance by capturing features across several scales, including spectrotemporal, phonemic, and some semantic information. Finally, we directly probed the SSL layers for acoustic and linguistic features, demonstrating that SSL models seem to recapitulate the speech processing hierarchy surprisingly well.

Overall, our results suggest an isomorphism between the representational depth of SSL models, the putative speech analysis hierarchy, and speech representations in human cortex. Coupled with our finding that SSL models are currently the best sound-based models of cortical speech processing, this suggests that further development and analysis of SSL models may reveal much about how speech is processed by the human brain.

\nocite{wolfTransformersStateoftheArtNatural2020} %
\bibliography{main}
\bibliographystyle{icml2022}

\newpage
\appendix
\onecolumn
\section{Voxel-wise encoding models of speech}

\subsection{Performance in speech ROIs}
\label{sec:supp_rois}

We show the encoding performance of each representation in two additional ROIs: Broca's area and sPMv.

\begin{figure*}[h!]
    \centering
    \includegraphics[scale=0.45]{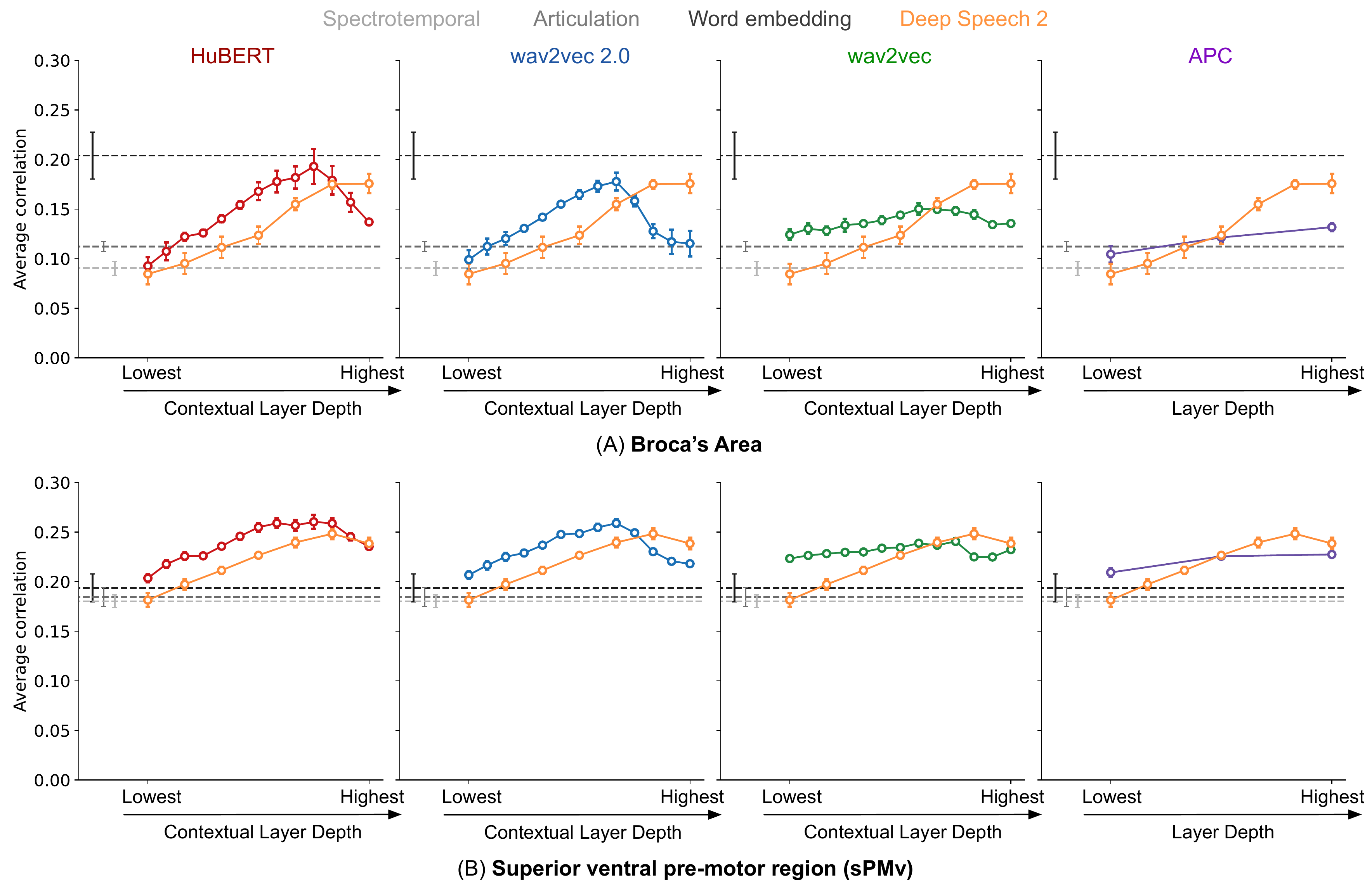}
    \caption{Average encoding performance across subjects ($N=6$), for every baseline and SSL model representation in two speech ROIs. As with \cref{fig:encoding-perf-layers}, error bars show subject-adjusted SEM.}
    \label{fig:supp_rois}
\end{figure*}

\section{Voxel-wise layer selectivity}
\label{sec:supp_layer-selectivity}

\begin{figure}[h!]
    \centering
    \begin{subfigure}{0.4\textwidth}
        \centering
        \includegraphics[width=\linewidth]{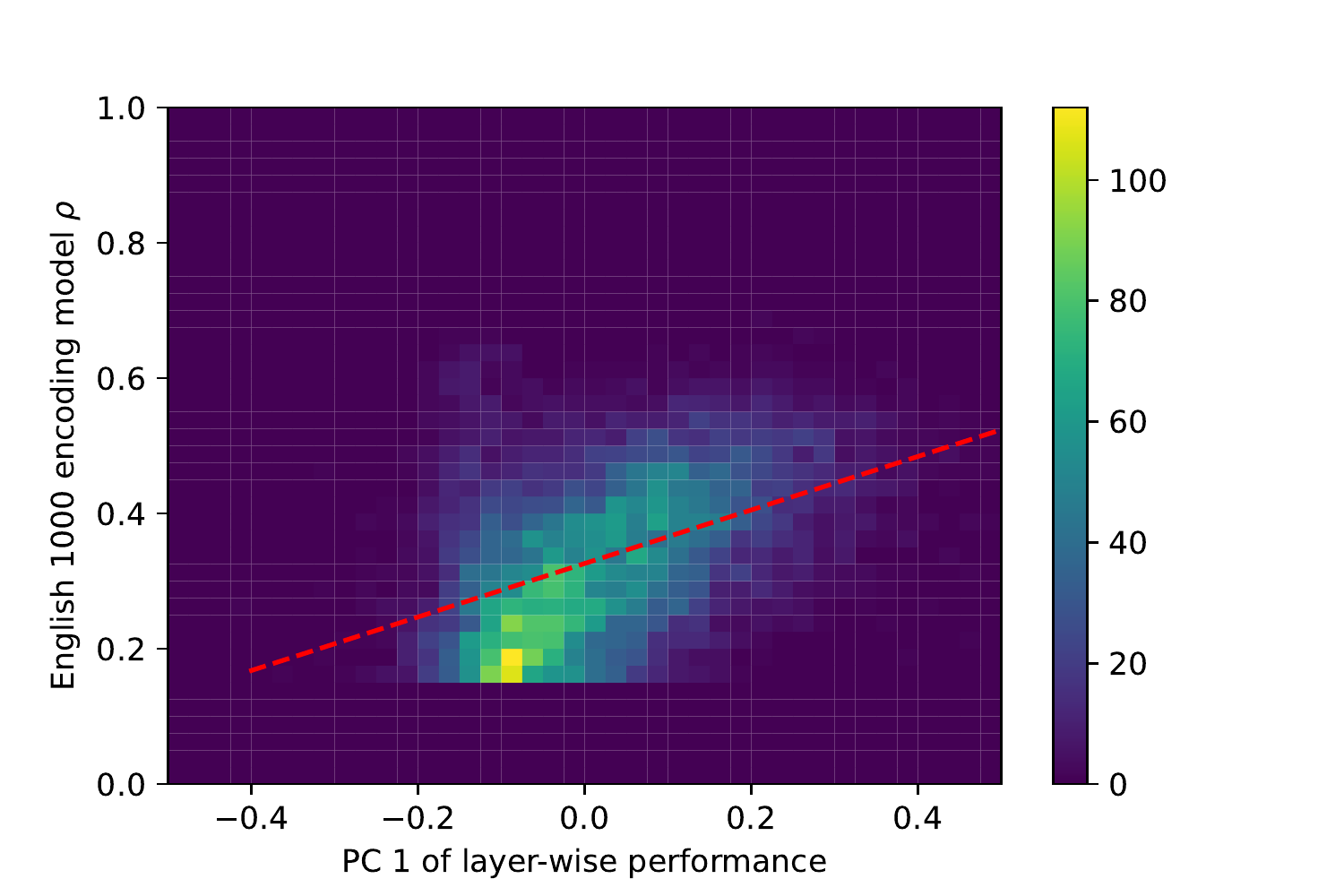}
        \caption{}
    \end{subfigure}
    \hfill
    \begin{subfigure}{0.4\textwidth}
        \centering
        \includegraphics[width=\linewidth]{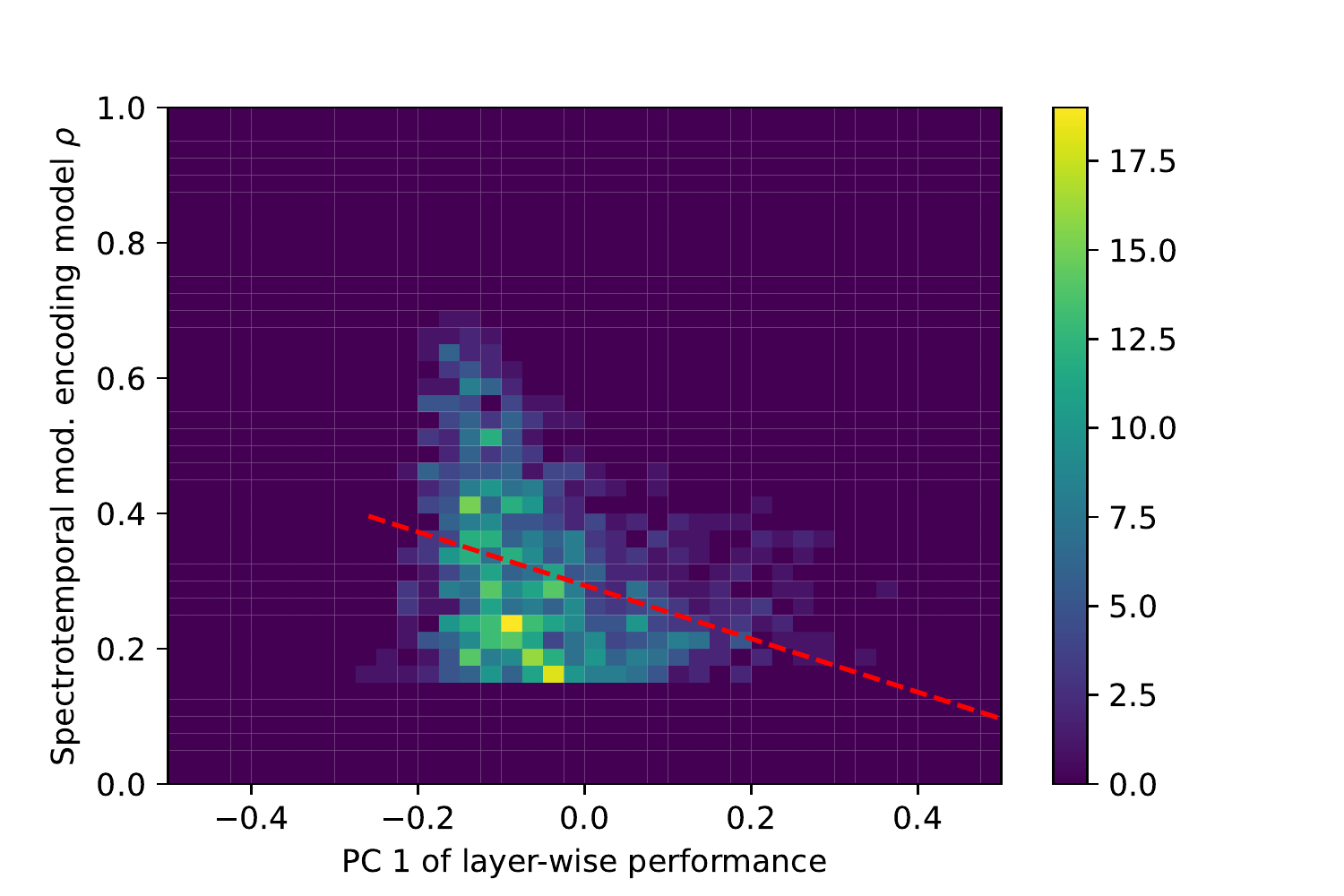}
        \caption{}
    \end{subfigure}
    \caption{2D histograms of voxels showing correlation between voxel-wise HuBERT layer selectivity (PC 1 in \cref{fig:layer-selectivity}) and two baselines. (a) Correlation of PC 1 and word embedding encoding performance across all of cortex are correlated ($\rho=0.449$), suggesting that PC 1 is positively correlated with semantics. (b) Correlation of PC 1 and spectrotemporal mod. encoding performance within voxels in AC. That these are negatively correlated ($\rho=-0.330$) gives further evidence that PC 1 is capturing higher-level processing.}
    \label{fig:supp_layer-selectivity}
\end{figure}

\section{Probing SSL model representations for linguistic structure}
\label{sec:supp_probing}

\begin{figure*}[h!]
    \centering
    \includegraphics[scale=0.5]{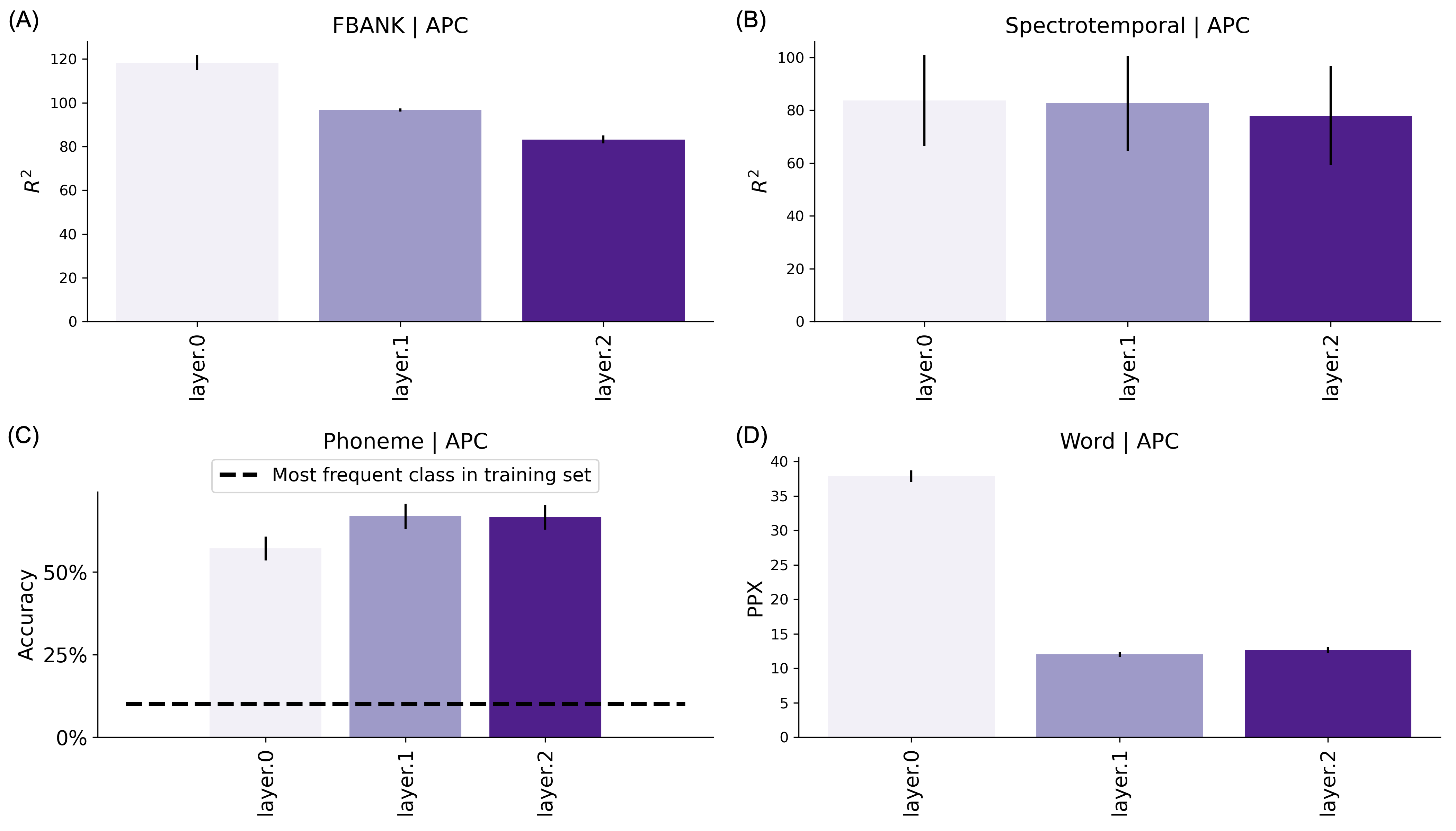}
    \caption{Probing results for APC.}
    \label{fig:probing_apc}
\end{figure*}

\begin{figure*}[h!]
    \centering
    \includegraphics[scale=0.5]{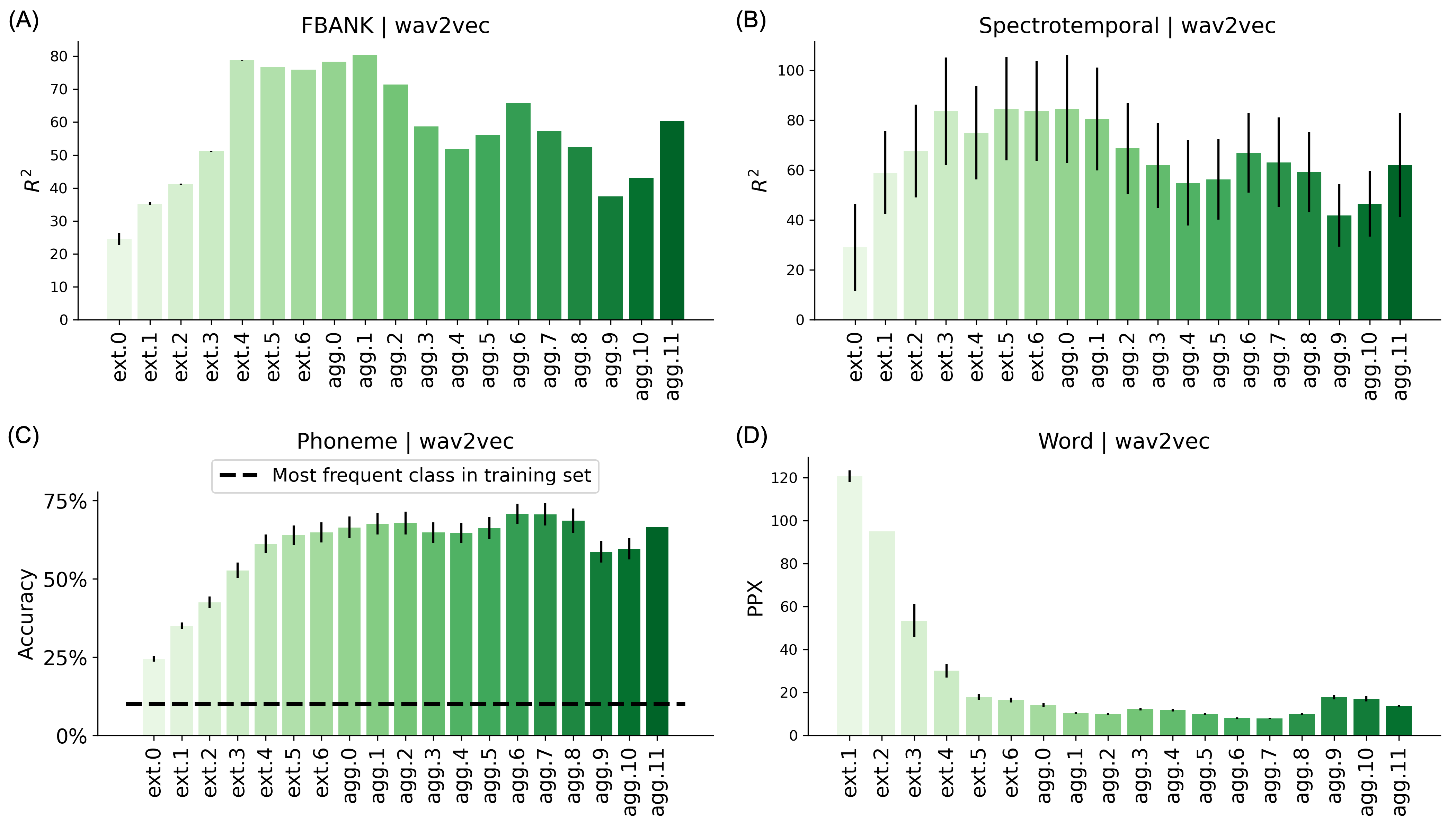}
    \caption{Probing results for wav2vec.}
    \label{fig:probing_wav2vec}
\end{figure*}

\begin{figure*}[h!]
    \centering
    \includegraphics[scale=0.5]{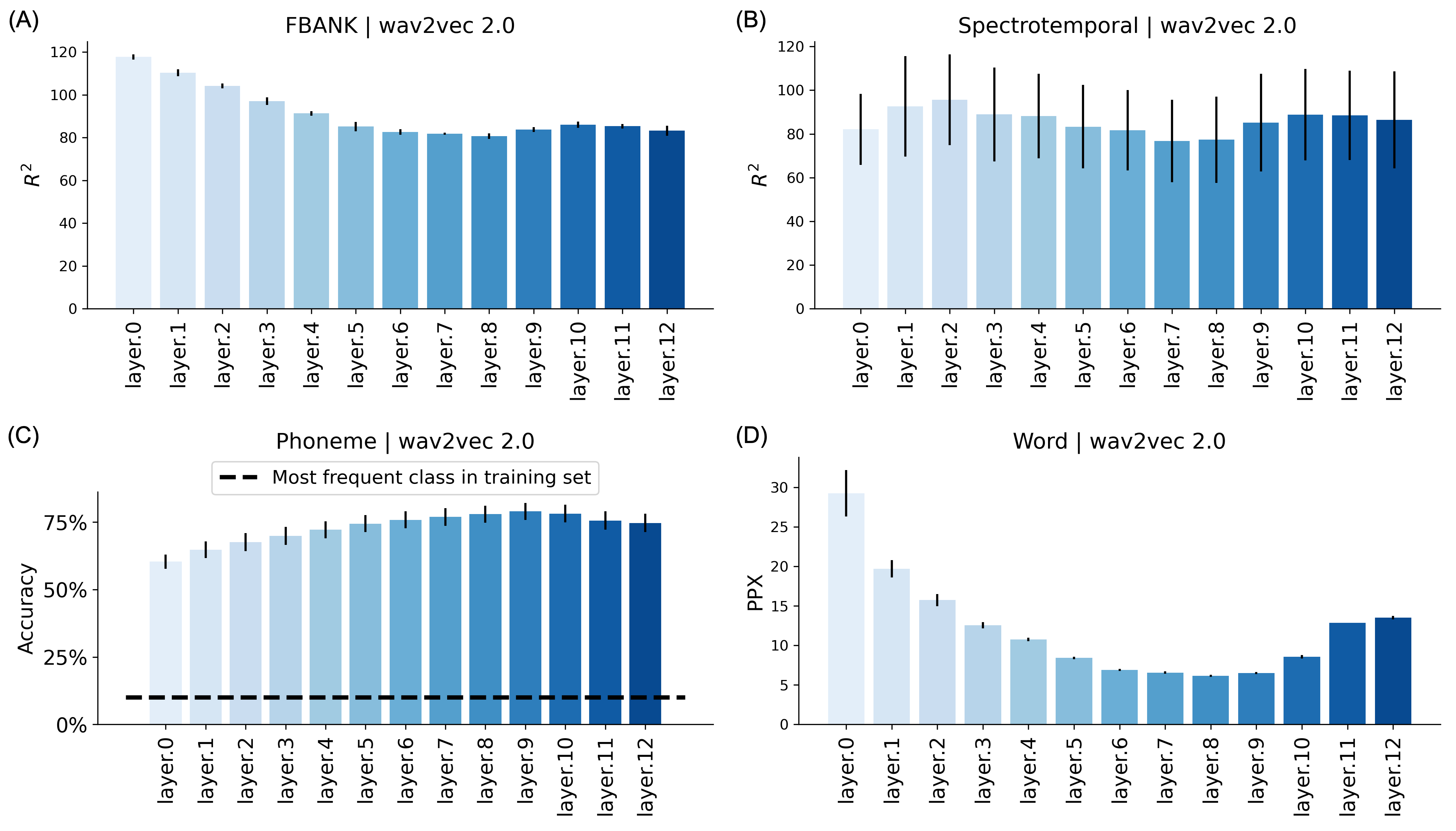}
    \caption{Probing results for wav2vec 2.0.}
    \label{fig:probing_wav2vec2}
\end{figure*}

\begin{figure*}[h!]
    \centering
    \includegraphics[scale=0.5]{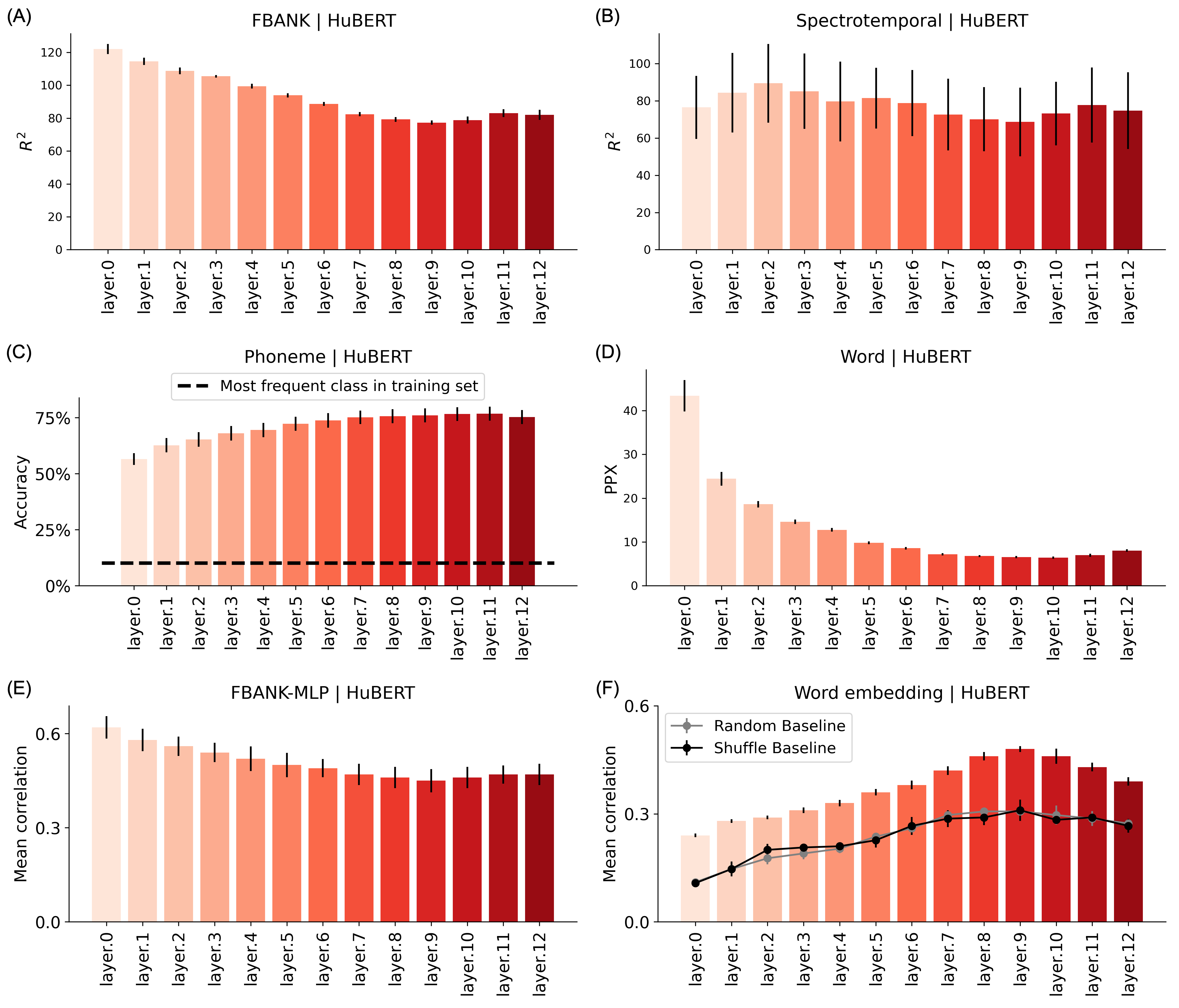}
    \caption{Probing results for HuBERT.}
    \label{fig:probing_hubert}
\end{figure*}

We visualize the full set of probing results for the four tasks (FBANK features, spectrotemporal features, phoneme identity, word identity) for each layer of the four SSL models (\cref{fig:probing_apc}-\cref{fig:probing_hubert}). Overall, the trends are consistent with the summaries reported in \cref{fig:probing} --- lower layers best predict low-level acoustic features, while the upper-middle layers are better at the phoneme and word tasks. In addition to the four probes, for HuBERT we also built MLP probes that predict the entire FBANK feature/GloVe embedding as opposed to regression probes that treat each feature in the representation independently. To do this, we trained a linear MLP with a single bottleneck layer (50-D). The model was trained on MSE loss and we applied L2 regularization to the output. Performance was measured as the linear correlation between true and predicted representation, averaged across the test set.

For the phoneme and word classification tasks, we developed a baseline that predicts the most frequent output category in the training set for all test set examples. All layers in the four SSL models beat the baseline. We do not visualize the word classification baseline perplexity since it is significantly worse than the highest layer PPX. For the word embedding MLP probe, we constructed two different baselines- randomly sampling embedding vectors from a normal distribution (``random'') and randomly shuffling the GloVe embeddings between words (``shuffle''). All layers in HuBERT beat the baseline with a considerable gap. Overall, the substantial differences in the performance of probing baselines and the SSL model layers suggest that the probes have high \textit{selectivity} \cite{hewittDesigningInterpretingProbes2019}.

\section{MRI acquistion, preprocessing, and experiment details}
\label{sec:supp_mri-acquisition}

\subsection{Participants}
All participants were healthy and had normal hearing, and normal or corrected-to-normal vision. To stabilize head motion during scanning sessions participants wore a personalized head case that precisely fit the shape of each participant's head. Anatomical data for subject S-02 were collected on a 3T Siemens TIM Trio scanner using a 32-channel Siemens volume coil at a different site. The same MP-RAGE sequence was used.
\subsection{Stimulus preparation and presentation}

Story stimuli were played over Sensimetrics S14 in-ear piezoelectric headphones. The audio for each story was filtered to correct for frequency response and phase errors induced by the headphones using calibration data provided by sensimetrics and custom Python code \footnote{Anonymized link.}%
All stimuli were played at 44.1 kHz using the pygame library in Python (https://www.pygame.org/news).

\subsection{Acquisition parameters}
Whole-brain MRI data was collected on a 3T Siemens Skyra scanner using a 64-channel Siemens volume coil. Functional MRI (fMRI) data were collected using a gradient echo EPI sequence, multi-band factor of 2. Scan parameters included repetition time (TR)=2.00s, echo time (TE)=30.8 ms, flip angle=71$^{\circ}$, voxel size=2.6mm\textsuperscript{3}, matrix size=84x84, field of view=220 mm. Anatomical MRI data were collected with a T1-weighted multi-echo MP-RAGE sequence with voxel size=1mm\textsuperscript{3}.
\subsection{Processing MRI data}
\paragraph{Functional data.} All functional data were motion corrected using the FMRIB Linear Image Registration Tool (FLIRT) from FSL 5.0. FLIRT was used to align all data to a template that was made from the average across the first functional run in the first story session for each subject. These automatic alignments were manually checked for accuracy.
\par Low frequency voxel response drift was identified using a 2nd order Savitzky-Golay filter %
with a 120 second window and then subtracted from the signal. To avoid onset artifacts and poor detrending performance near each end of the scan, responses were trimmed by removing 20 seconds (10 volumes) at the beginning and end of each scan, which removed the 10-second silent period and the first and last 10 seconds of each story. The mean response for each voxel was subtracted and the remaining response was scaled to have unit variance.

To account for physiological and behavioral noise, encoding models for one analysis (layer selectivity) were fit with responses that had been corrected for nuisance regressors that accounted for closed-eye eye movements and head motion. This has little effect on the overall model performance, but accounts for spurious eye movement and stimulus-related responses in visual cortex.

\paragraph{Anatomical data.}  Cortical surface meshes were generated from the T1-weighted anatomical scans using FreeSurfer \cite{daleCorticalSurfaceBasedAnalysis1999}. Before surface reconstruction, anatomical surface segmentations were hand-checked and corrected. Blender was used to remove the corpus callosum and make relaxation cuts for flattening. Functional images were aligned to the cortical surface using boundary based registration (BBR) in FSL. %
These alignments were manually checked for accuracy and adjustments were made as necessary. Flat maps were created by projecting the values for each voxel onto the cortical surface using the ``nearest'' scheme in pycortex \cite{gaoPycortexInteractiveSurface2015}.
\subsection{Defining regions of interest (ROIs)}
Known regions of interest (ROIs) were localized separately in each participant. Three different tasks were used to define ROIs; a visual category localizer, an auditory cortex localizer, and a motor localizer. 

For the visual category localizer, data were collected in six 4.5 minute scans consisting of 16 blocks of 16 seconds each. During each block 20 images of either places, faces, bodies, household objects, or spatially scrambled objects were displayed. participants were asked to pay attention to the same image being presented twice in a row. The corresponding ROIs defined in the cerebral cortex with this localizer were the fusiform face area (FFA), occipital face area (OFA), extrastriate body area (EBA), parahippocampal place area (PPA), and occipital place area (OPA).

The motor localizer data were was collected during 2 identical 10-minute scans. The participant was cued to perform six different tasks in a random order in 20 second blocks. The cues were ``hand'', ``foot'', ``mouth'', ``speak'', saccade, and ``rest'' presented as a word at the center of the screen, except for the saccade cue which was presented as an array of dots. For the ``hand'' cue, participants were instructed to make small finger-drumming movements for the entirety of the cue display. For the ``foot'' cue, participants were instructed to make small foot and toe movements. For the ``mouth'' cue, participants were instructed to make small vocalizations that were nonsense syllables such as balabalabala. For the ``speak'' cue, participants were instructed to self-generate a narrative without vocalization. For the saccade cue, participants were instructed to make frequent saccades across the display screen for the duration of the task. 

Weight maps for the motor areas were used to define primary motor and somatosensory areas for the hands, feet, and mouth; supplemental motor areas for the hands and feet, secondary somatosensory areas for the hands, feet, and mouth, and the ventral premotor hand area. The weight map for the saccade responses was used to define the frontal eye fields and intraparietal sulcus visual areas. The weight map for speech was used to define Broca’s area and the superior ventral premotor (sPMv) speech area. 

Auditory cortex localizer data were collected in one 10 minute scan. The participant listened to 10 repeats of a 1-minute auditory stimulus containing 20 seconds of music (Arcade Fire), speech (Ira Glass, This American Life), and natural sound (a babbling brook). To determine whether a voxel was responsive to auditory stimulus, the repeatability of the voxel response across the 10 repeats was calculated using an F-statistic. This map was used to define the auditory cortex (AC).

\end{document}